%% file: acl_latex.tex
\pdfoutput=1

\documentclass[11pt]{article}

\usepackage[preprint]{acl}

\usepackage{times}
\usepackage{latexsym}

\usepackage[T1]{fontenc}

\usepackage[utf8]{inputenc}

\usepackage{microtype}

\usepackage{inconsolata}

\usepackage{graphicx}

\usepackage{amssymb}
\usepackage{amsmath}

\usepackage{fontawesome}

\usepackage{algorithm}
\usepackage{algorithmic}
\usepackage{xcolor} %

\usepackage{multirow}

\usepackage{hyperref}
\makeatletter
\def\UrlAlphabet{%
      \do\a\do\b\do\c\do\d\do\e\do\f\do\g\do\h\do\i\do\j%
      \do\k\do\l\do\m\do\n\do\o\do\p\do\q\do\r\do\s\do\t%
      \do\u\do\v\do\w\do\x\do\y\do\z\do\A\do\B\do\C\do\D%
      \do\E\do\F\do\G\do\H\do\I\do\J\do\K\do\L\do\M\do\N%
      \do\O\do\P\do\Q\do\R\do\S\do\T\do\U\do\V\do\W\do\X%
      \do\Y\do\Z}
\def\UrlDigits{\do\1\do\2\do\3\do\4\do\5\do\6\do\7\do\8\do\9\do\0}
\g@addto@macro{\UrlBreaks}{\UrlOrds}
\g@addto@macro{\UrlBreaks}{\UrlAlphabet}
\g@addto@macro{\UrlBreaks}{\UrlDigits}
\makeatother

\input{latex/formula_notation}

\title{
EliteKV: Scalable KV Cache Compression \\via RoPE Frequency Selection and Joint Low-Rank Projection
}

\author{
Yuhao Zhou\thanks{~Equal contribution.}, 
Sirui Song\footnotemark[1], 
Boyang Liu, \\ \bf
Zhiheng Xi,
Senjie Jin, 
Xiaoran Fan, 
Zhihao Zhang, 
Wei Li, 
Xuanjing Huang\thanks{~Corresponding authors.}
\\ 
Fudan University
\\
\texttt{
\{zhouyh24,srsong23\}@m.fudan.edu.cn, xjhuang@fudan.edu.cn
}
}

\begin{document}
\maketitle

\input{sections/0_abstract}
\input{sections/1_intro}

\input{sections/2_background}

\input{sections/3_framework}

\input{sections/4_experiments}

\input{sections/5_related}

\input{sections/6_conclusion}
\input{sections/8_limitations}

\bibliography{anthology,custom}

\appendix

\input{sections/7_appendix}

\end{document}

%% file: latex/formula_notation.tex
\newcommand{\timestep}{t}
\newcommand{\qlen}{m}
\newcommand{\klen}{n}
\newcommand{\head}{h}
\newcommand{\numheads}{n_{h}}
\newcommand{\hiddensize}{d}
\newcommand{\headdim}{d_{h}}
\newcommand{\loopi}{i}
\newcommand{\loopj}{j}

\newcommand{\ropemethod}{\textit{RoPElite}}
\newcommand{\ourmethod}{\textit{EliteKV}}

%% file: sections/0_abstract.tex
\begin{abstract}

Rotary Position Embedding (RoPE) enables each attention head to capture multi-frequency information along the sequence dimension and is widely applied in foundation models. 
However, the nonlinearity introduced by RoPE complicates optimization of the key state in the Key-Value (KV) cache for RoPE-based attention. 
Existing KV cache compression methods typically store key state before rotation and apply the transformation during decoding, introducing additional computational overhead. 
This paper introduces \ourmethod{}, a flexible modification framework for RoPE-based models supporting variable KV cache compression ratios.
\ourmethod{} first identifies the intrinsic frequency preference of each head using \ropemethod{}, selectively restoring linearity to certain dimensions of key within attention computation. 
Building on this, joint low-rank compression of key and value enables partial cache sharing. 
Experimental results show that with minimal uptraining on only $0.6\%$ of the original training data, RoPE-based models achieve a $75\%$ reduction in KV cache size while preserving performance within a negligible margin.
Furthermore, \ourmethod{} consistently performs well across models of different scales within the same family.
\footnote{~Codes are publicly available at \href{https://github.com/CiaranZhou/EliteKV}{\faGithub{} CiaranZhou/EliteKV}.}

\end{abstract}

%% file: sections/1_intro.tex
\input{figures_latex/banner}

\section{Introduction}

Large language foundation models demonstrates remarkable capabilities in solving a wide range of problems in the language domain \citep{DBLP:journals/corr/abs-2412-16720, DBLP:journals/corr/abs-2307-09288, DBLP:journals/corr/abs-2501-12948}. 
To accelerate inference, a simple yet effective technique is to cache the key and value (KV cache) in the attention mechanism \citep{DBLP:conf/nips/VaswaniSPUJGKP17} to avoid redundant computations. 
However, the size of the KV cache grows linearly with the number of decoding steps \citep{DBLP:conf/mlsys/PopeDCDBHXAD23, DBLP:journals/corr/abs-2405-08944}, which becomes a limitation for real-world, long-text, and real-time applications \citep{DBLP:journals/corr/abs-2412-19442}.
Therefore, compressing the KV cache becomes a key issue for improving inference efficiency.

A mainstream approach to compressing the KV cache is to perform low-rank decomposition on the key and value projection matrices, which allows replacing the cached content with lower-dimensional intermediate states \citep{DBLP:journals/corr/abs-2412-06867, wang2025svdllm}.
However, with the widespread use of rotary position embedding (\citealp[RoPE]{DBLP:journals/ijon/SuALPBL24}), which have nonlinear properties, the application of such methods becomes challenging. 
The nonlinear nature of RoPE prevents caching the rotated intermediate states with lower dimensions, limiting the cache to storing the pre-rotated states \citep{chang2025palu}. 
When applying RoPE again to all cache entries during usage, additional computation is introduced during decoding, which contradicts the original design intention of the KV cache \citep{DBLP:journals/corr/abs-2410-03111, DBLP:journals/corr/abs-2410-21465, chang2025palu}.

This work describes \ourmethod{}, a straightforward method that bypasses the constraints imposed by the non-linear nature of RoPE and allows for the compression of KV cache to any specified proportion of the original size.
\ourmethod{} aims to eliminate RoPE in certain dimensions of the attention computation flow, enabling the compression of the cache formed by those dimensions. 
Specifically, first, based on the varying frequency preferences of each attention head, the combination of dimensions in the rotation that each attention head relies on the most is identified. 
Then, a joint low-rank decomposition (J-LRD) is performed on the projection matrices of the parts of the key without RoPE and the entire value for all heads in the same layer, enabling them to be decomposed into a shared down-projection matrix and corresponding up-projection matrix.
The resulting modified attention computation flow, as shown in Figure \ref{fig:banner}. 
Each head only selects the frequency information most relevant to it in the sequence dimension for RoPE which call \ropemethod{}.
Beyond that, the remaining parts of the key state without RoPE and the entire value state are represented by a shared low-dimensional intermediate state.

Experimental results demonstrate the superiority of \ourmethod{} as a method that modifies the model structure to achieve KV cache compression. 
It enables reducing the KV cache to just $25\%$ of the original size while maintaining comparable performance. 
Additionally, when the KV cache is reduced to $12.5\%$, the performance is on par with GQA at $50\%$.
Ablation studies further validate the effectiveness of \ropemethod{} in searching for the optimal rotation dimension combination for each attention head. 
Moreover, J-LRD outperforms traditional low-rank decomposition methods, which treat each weight matrix separately, in terms of KV cache size reduction. 
As the model parameters scale, \ourmethod{} shows predictable performance loss across different model sizes within the same family, highlighting its scalability and efficiency in maintaining performance.

%% file: figures_latex/banner.tex
\begin{figure*}[tb]
\centering
\includegraphics[width=.95\linewidth]{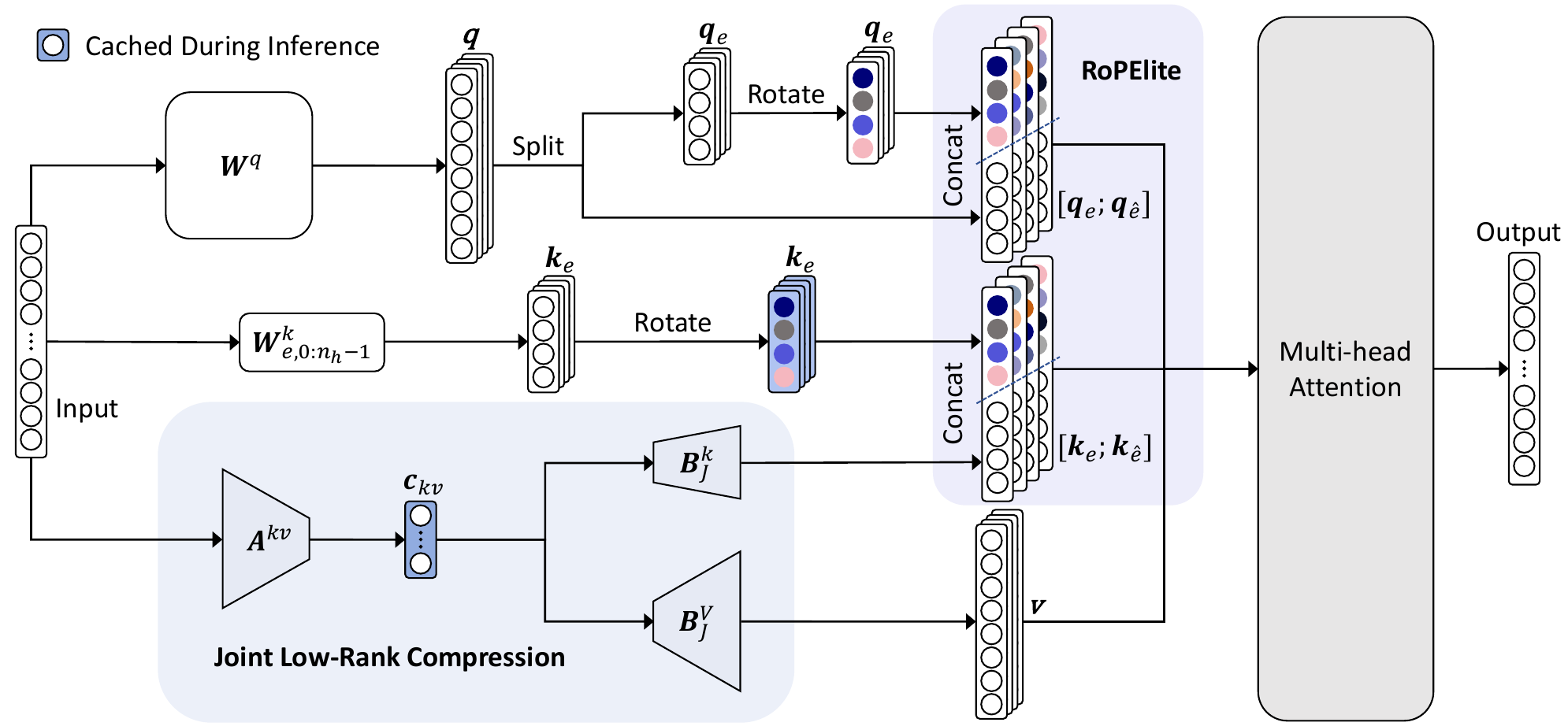}
\caption{
The attention computation flow after applying \ourmethod{}. 
The upper part illustrates \ropemethod{}, where each attention head focuses only on its most important frequency along the sequence dimension. The lower part shows the joint low-rank projection, where the K and V states are represented by a shared cache.
The different colored fillings in the elements represent 2D chunks attending to different frequencies.
}
\label{fig:banner}
\end{figure*}

%% file: sections/2_background.tex
\section{Background}

\subsection{Attention Mechanism and KV Cache}

The attention mechanism is a core component of the Transformer model, which measures the importance of different tokens, allowing the model to focus on different parts of the input sequence \citep{DBLP:conf/nips/VaswaniSPUJGKP17}. 
Given an input $\mathbf{x}_{\timestep} \in \mathbb{R}^{\hiddensize}$ at time step $\timestep$, it is projected onto query, key, and value for each attention head $\head$ using the weight matrices $\mathbf{W}^{q}, \mathbf{W}^{k}, \mathbf{W}^{v} \in \mathbb{R}^{\hiddensize \times \headdim}$ 
\footnote{Processes refer to a single attention head unless stated otherwise, with the subscript $\head$ omitted for simplicity.}, 
where $\hiddensize$ represents the embedding dimension of each token and $\headdim$ denotes the dimension of each attention head.
The process is formulated as follows:
$\mathbf{q}_{\timestep} = \mathbf{x}_{\timestep} \mathbf{W}^{q}, 
\mathbf{k}_{\timestep} = \mathbf{x}_{\timestep} \mathbf{W}^{k} \text{ and }
\mathbf{v}_{\timestep} = \mathbf{x}_{\timestep} \mathbf{W}^{v}.$
Then the attention mechanism is as follows:
\begin{equation*}
\begin{aligned}
    \mathbf{K}_{\timestep} = \sum_{\loopj=1}^{\timestep}\mathbf{k}_{\loopj},
    \hspace{0.5em}
    \mathbf{V}_{\timestep} = \sum_{\loopj=1}^{\timestep}\mathbf{v}_{\loopj},&
    \hspace{0.5em}
    \mathbf{s}_{\timestep} = \mathbf{q}_{\timestep}\mathbf{K}_{\timestep}^{\top},\\
    \mathbf{p}_{\timestep}=\mathrm{Softmax}\left(\frac{\mathbf{s}_{\timestep}}{\sqrt{\headdim}}\right),&
    \hspace{0.5em}
    \mathbf{o}_{\timestep}=\mathbf{p}_{\timestep}\mathbf{V}_{\timestep}.
\end{aligned}
\end{equation*}

The autoregressive nature of decoder-only models allows for the reuse of key-value pairs cached from previous tokens (KV cache), reducing the computational cost during token-by-token decoding. At each decoding step $\timestep$, the $\mathbf{K}_{\timestep}$ and $\mathbf{V}_{\timestep}$ are obtained by concatenating the cached $\mathbf{K}_{\timestep-1}$ and $\mathbf{V}_{\timestep-1}$ with the newly computed $\mathbf{k}_{\timestep}$ and $\mathbf{v}_{\timestep}$.

\subsection{Rotary Position Embedding (RoPE)}

RoPE \citep{DBLP:journals/ijon/SuALPBL24} is a widely used method for incorporating positional information into Transformer models. It divides the dimensions of the query and key for each attention head into several 2D pairs, with each pair capturing information at different frequencies $\theta$ along the sequence dimension, thereby enabling the attention score $\mathbf{s}$ to carry positional information as shown by:
\begin{subequations}
\begin{align}
{\mathbf{s}_{\qlen}}_{\left[\klen\right]}
&=\sum_{\loopi \in \mathcal{I}}
\left( \mathbf{q}_{\qlen,\loopi} \mathbf{R}(\qlen \theta_{\loopi}) \right)
\left( \mathbf{k}_{\klen,\loopi} \mathbf{R}(\klen \theta_{\loopi}) \right)^{\top} \label{eq:rope_absolute} \\
&=\sum_{\loopi \in \mathcal{I}}
\mathbf{q}_{\qlen,\loopi}
\mathbf{R}\left((\qlen - \klen) \theta_{\loopi}\right)
\mathbf{k}_{\klen,\loopi}^{\top} \label{eq:rope_relative}
\end{align}
\end{subequations}
where $\mathbf{R}$ is a 2D rotation matrix,
\begin{equation*}
    \mathcal{I} = \{ [2i:2i+1] \mid i = 0, 1, 2, \dots, \frac{\headdim}{2} - 1 \}.
\end{equation*}
For RoPE-based models, at the $\timestep$-th decoding step, 
\begin{equation*}
\sum_{\loopi \in \mathcal{I}}\mathbf{k}_{\timestep,\loopi} \mathbf{R}(\timestep \theta_{\loopi}) + \mathbf{v}_{\timestep}
\end{equation*}
are incorporated into the KV cache.

\subsection{Singular Value Decomposition (SVD)}

SVD is a method widely used for dimensionality reduction and matrix approximation.
Given a matrix $\mathbf{M} \in \mathbb{R}^{m \times n}$, the SVD decomposes it as $\mathbf{U} \mathbf{\Sigma} \mathbf{V}^{\top}$, where $\mathbf{U} \in \mathbb{R}^{m \times m}$ and $\mathbf{V} \in \mathbb{R}^{n \times n}$ are orthogonal matrices, and $\mathbf{\Sigma} \in \mathbb{R}^{m \times n}$ is a diagonal matrix with non-negative elements arranged in descending order along the diagonal.
Let $\mathbf{A} = \mathbf{U}$, and $\mathbf{B} = \mathbf{\Sigma} \mathbf{V}^{\top}$, then the optimal rank-$r$ approximation of the matrix $\mathbf{M}$ is $\mathbf{A}_{[:,:r]} \mathbf{B}_{[:r,:]}$, which changes the storage requirement to $\frac{mr + rn}{mn}$ times the original.

%% file: sections/3_framework.tex
\input{figures_latex/top-r}

\section{Framework of \ourmethod{}}

\subsection{\ropemethod{}: Retain the Frequency Most Attended to by Each Attention Head}

Although different heads follow the same set of frequencies to compute attention scores during the model structure design phase \citep{DBLP:journals/corr/abs-2407-21783, DBLP:journals/corr/abs-2412-15115, DBLP:journals/corr/abs-2412-19437}, 
some prior works observe that different attention heads exhibit varying sensitivity to different frequency information when computing attention scores \citep{hong-etal-2024-token,barbero2025round}, implying that the chunks in $\mathcal{I}$ have different impacts on $\mathbf{s}$. We refer to the most influential chunks as \textit{\textbf{e}lite chunks}, and the top-$r$ chunks formed by these \textit{\textbf{e}lite chunks} are denoted as $\mathcal{I}_{r}^{e}$.

\input{algorithms/RoPElite}

Finding $\mathcal{I}_{r}^{e}$ for each attention head is challenging, as there are $\binom{\frac{h_d}{2}}{r}$ ways to choose $r$ elements from $\mathcal{I}$, and the model has $l$ layers with $n_h$ heads per layer. 
Based on this, we propose \ropemethod{}, which uses a greedy algorithm to find $\mathcal{I}_{r}^{e}$ for each attention head, as shown in Algorithm~\ref{alg:RoPElite}.
It fully leverages parallelism, enabling the traversal of $l$ layers and $n_h$ heads in a single forward pass, reducing the time complexity to $O(rh_d)$, independent of $l$ and $n_h$ (detailed analysis in Appendix \ref{sec:appendix time comsumption}).

As an example, for the model LLaMA2-7B, where each head is trained on $\mathcal{I}$, the $\mathcal{I}_{8}^{e}$ distributions across different layers and heads are visualized in Figure \ref{fig:top-r} (more detailed plots are in Appendix \ref{sec:appendix pattern}, Figure \ref{fig:appendix top-r}).
It is clear that different attention heads focus on different frequency patterns across the sequence dimension. 
Taking $\mathcal{I}_{8}^{e}$ as an example, the frequency dependencies of attention heads can be categorized as follows (L for layer and H for head): primarily high-frequency (L0H22, L0H31), primarily low-frequency (L0H8, L23H8, L31H22), mixed extreme frequencies (L0H0, L0H26), and predominantly mid-frequency (most heads).
Notably, high-frequency information is predominantly captured by heads in shallow layers.

\input{figures_latex/performance_of_top-r}

After applying \ropemethod{}, the calculation process of the attention scores changes from the Equation~\ref{eq:rope_relative} to
\begin{equation*}\resizebox{\linewidth}{!}{$
{\mathbf{s}_{\qlen}}_{\left[\klen\right]}
= {\displaystyle\sum_{\loopi \in \mathcal{I}_{r}^{e}}} \mathbf{q}_{\qlen,\loopi} \mathbf{R}\left((\qlen - \klen) \theta_{\loopi}\right) \mathbf{k}_{\klen,\loopi}^{\top} + {\displaystyle\sum_{\loopi \in \mathcal{I} \setminus \mathcal{I}_{r}^{e}}} \mathbf{q}_{\qlen,\loopi} \mathbf{k}_{\klen,\loopi}^{\top} .
$}\end{equation*}
Such modifications inevitably require uptraining the model.
For different $\mathcal{I}_{r}^{e}$, the model's performance varies with training, as shown in Figure \ref{fig:performance of top-r}.
It can be observed that using only a small proportion of tokens, relative to the model's original training tokens, during the uptraining phase can restore or even surpass the original performance.
This suggests that retaining only the dimensions of high importance at the head level in a fully RoPE-trained model can maintain near full-frequency performance at the model level.
A similar phenomenon was also observed by \citet{lu-etal-2024-longheads}.
This implies that dimensions with lower importance may act as Gaussian noise in the calculation of attention scores.

For \ropemethod{}-based models, at the $\timestep$-th decoding step, 
\begin{equation*}
\sum_{\loopi \in \mathcal{I}_{r}^{e}}\mathbf{k}_{\timestep,\loopi} \mathbf{R}(\timestep \theta_{\loopi})+\sum_{\loopi \in \mathcal{I} \setminus \mathcal{I}_{r}^{e}}\mathbf{k}_{\timestep,\loopi} + \mathbf{v}_{\timestep}
\end{equation*}
are incorporated into the KV cache, which is consistent in size with that of RoPE-based models.
The detailed workflow of \ropemethod{} in the attention mechanism is shown on the upper part of Figure \ref{fig:banner}.

\input{figures_latex/S-LRD_and_J-LRD}

\subsection{Low-Rank Compression for KV Cache}
\label{sec:LRD}

Optimizing the KV cache by low-rank decomposition of $\mathbf{W}^{k}$ and $\mathbf{W}^{v}$ in RoPE-based models is challenging. 
RoPE employs absolute position encoding (Equation \ref{eq:rope_absolute}) to achieve relative position encoding (Equation \ref{eq:rope_relative}), which requires each head's K cache to store $\headdim = |\mathcal{I}| \times 2$ elements per token, otherwise additional computation is needed during decoding (see Section 5). 
The use of \ropemethod{} alleviates this constraint by enabling partial compression of $\mathbf{k}$, bypassing the difficult-to-handle part of the RoPE cache and allowing the optimization of the KV cache size through $\sum_{\loopi \in \mathcal{I} \setminus \mathcal{I}_{r}^{e}}\mathbf{k}_{\timestep,\loopi}$ and $\mathbf{v}_{\timestep}$.

Let $\mathbf{W}^{k}_{e, h} \in \mathbb{R}^{\hiddensize \times (2r)}$ denotes the portion of the key projection matrix for each attention head $\head$ that is responsible for $\mathcal{I}_{r}^{e}$, while $\mathbf{W}^{k}_{\hat{e}, h} \in \mathbb{R}^{\hiddensize \times (\headdim-2r)}$ represents the remaining part.
To fully leverage the shared information across multiple heads, we consider performing low-rank decomposition on the projection matrix formed by all attention heads, rather than decomposing it on a per-head basis. 
Specifically, we handle the following combined projection matrices:
\begin{equation}
\begin{aligned}
\mathbf{W}^{k}_{\hat{e},0:\numheads-1} &= \left[ \mathbf{W}^{k}_{\hat{e}, 0}, \mathbf{W}^{k}_{\hat{e}, 1} \dots \mathbf{W}^{k}_{\hat{e}, \numheads-1} \right],\\
\mathbf{W}^{v}_{0:\numheads-1} &= \left[ \mathbf{W}^{v}_{0}, \mathbf{W}^{v}_{1} \dots \mathbf{W}^{v}_{\numheads-1} \right].
\end{aligned}
\end{equation}
Depending on whether the K and V caches are optimized together, we explore \textit{separated low-rank decomposition (S-LRD)} and \textit{joint low-rank decomposition (J-LRD)}.

\noindent\textbf{Separated Low-Rank Decomposition}
It is common to apply SVD for low-rank factorization of each weight matrix in the model \citep{DBLP:journals/corr/abs-2410-03111, chang2025palu, wang2025svdllm}, as it effectively reduces the model's storage cost and the intermediate dimensionality.
S-LRD allows for the customization of rank selection for each matrix, leading to more precise approximations, which in turn minimizes performance degradation.
S-LRD approximates the two matrices separately as:
\begin{equation*}
\mathbf{W}^{k}_{\hat{e},0:\numheads-1} \approx \mathbf{A}^{k}\mathbf{B}^{k}_{S},\hspace{0.5em}
\mathbf{W}^{v}_{0:\numheads-1} \approx \mathbf{A}^{v}\mathbf{B}^{v}_{S},
\end{equation*}
where
\begin{equation*}
\begin{aligned}
\mathbf{A}^{k} \in \mathbb{R}^{\hiddensize \times d_{c_k}}&,\hspace{0.5em} 
\mathbf{B}^{k}_{S} \in \mathbb{R}^{d_{c_k} \times (\headdim\numheads - 2r\numheads)}, \\
\mathbf{A}^{v} \in \mathbb{R}^{\hiddensize \times d_{c_v}}&,\hspace{0.5em} 
\mathbf{B}^{v}_{S} \in \mathbb{R}^{d_{c_v} \times (\headdim\numheads)}.
\end{aligned}
\end{equation*}
The storage cost of \ropemethod{} with S-LRD is given by:
\begin{equation*}
2r\numheads\hiddensize + d_{c_k}(\hiddensize + \headdim\numheads - 2r\numheads) + d_{c_v}(\hiddensize + \headdim\numheads),
\end{equation*}
which simplifies to:
\begin{equation*}
(2d_{c_k} + 2d_{c_v} + 2r\numheads)\hiddensize - 2d_{c_k} r \numheads,
\end{equation*}
in the case of MHA models, where the structural design typically follows $\hiddensize = \headdim \numheads$. 
The KV cache consumption per token per layer is $2r\numheads + d_{c_k} + d_{c_v}$.

\noindent\textbf{Joint Low-Rank Decomposition}
From the perspective of sharing the KV cache, we propose J-LRD, which performs low-rank decomposition while simultaneously considering both K and V, leveraging the shared information between the two projection matrices.
In J-LRD, the concatenated weight matrices are factorized as
\begin{equation*}
\mathbf{W}^{kv} = \left[\mathbf{W}^{k}_{\hat{e},0:\numheads-1}, \mathbf{W}^{v}_{0:\numheads-1}\right] \approx \mathbf{A}^{kv} \mathbf{B}^{kv},
\end{equation*}
where 
\begin{equation*}
\mathbf{A}^{kv} \in \mathbb{R}^{\hiddensize \times d_{c_{kv}}},\hspace{0.5em}
\mathbf{B}^{kv} \in \mathbb{R}^{d_{c_{kv}} \times (2\headdim\numheads - 2r\numheads)}.
\end{equation*}
The matrix used for up-projection is then split into:
\begin{equation*}
\mathbf{B}^{kv} = [\mathbf{B}^{k}_{J}, \mathbf{B}^{v}_{J}],
\end{equation*}
where
\begin{equation*}
\mathbf{B}^{k}_{J} \in \mathbb{R}^{d_{c_{kv}} \times (\headdim\numheads - 2r\numheads)},
\hspace{0.5em}
\mathbf{B}^{v}_{J} \in \mathbb{R}^{d_{c_{kv}} \times (\headdim\numheads)}.
\end{equation*}
Under this factorization, the storage cost of \ropemethod{} with J-LRD is expressed as:
\begin{equation*}
2r\numheads\hiddensize + d_{c_{kv}}(\hiddensize + 2\headdim\numheads - 2r\numheads).
\end{equation*}
By simplifying the expression under the common MHA assumption, this reduces to:
\begin{equation*}
2r\numheads\hiddensize + 3d_{c_{kv}}\hiddensize - 2d_{c_{kv}}r\numheads.
\end{equation*}
Furthermore, the KV cache requirement per token per layer in this setting is $2r\numheads + d_{c_{kv}}$.

The workflows of S-LRD and J-LRD are shown in Figure \ref{fig:different LRD}, with the matrix decomposition process depicted on the left side of the figure, and the role of the resulting matrices in the new attention computation flow shown on the right side.  
J-LRD is chosen as the final component, and a detailed analysis of the experiments is provided in Section \ref{sec:ablation LRD}.

Finally, the framework of \ropemethod{} with J-LRD is illustrated in Figure \ref{fig:banner}.  
From the perspective of KV cache, the approach first eliminates less critical RoPE dimensions in the original model, creating space for the remaining KV cache components to be projected into a low-rank latent space. 
Notably, this eliminates the need to reapply the rotation to the keys in the cache during each decoding step.
Next, leveraging the shared information between key and value, the projection matrices of both are jointly decomposed, enabling key-value pairs to be stored in a shared cache. 
This design is both parameter-efficient and KV cache-efficient.  
Finally, the optimal dimension configurations are selected based on simple filtering principles. 
The model is then uptrained from its unmodified state using the selected dimensions, resulting in a KV cache-efficient model.

%% file: figures_latex/top-r.tex
\begin{figure*}[tb]
\centering
\includegraphics[width=\linewidth]{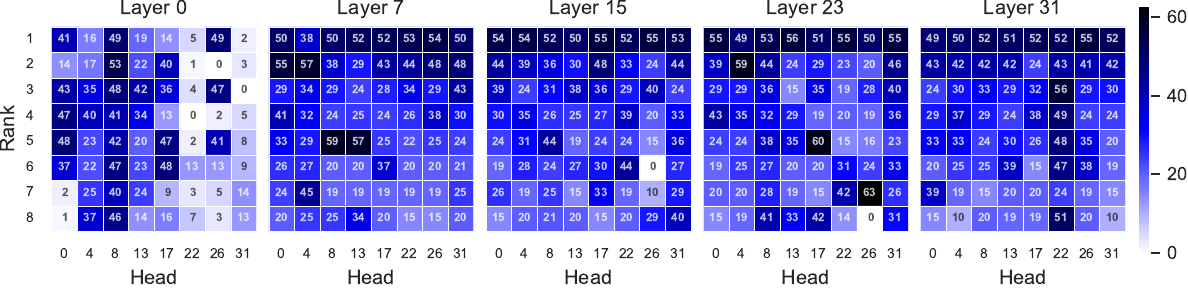}
\caption{
Top-$8$ chunks of different attention heads in different layers. 
Frequency preference patterns of different attention heads across layers in the LLaMA2-7B model. 
Numbers increase from high to low frequencies.
}
\label{fig:top-r}
\end{figure*}

%% file: algorithms/RoPElite.tex
\begin{algorithm}[tb]
    \caption{RoPElite}
    \begin{algorithmic}
        \REQUIRE head dimension $\headdim$, collection of Rope pairs $\mathcal{I}$, number of elite chunks $r$
        \ENSURE elite chunks $e$
        \STATE Initialize $e \gets \emptyset$ as an empty list
    
        \FOR{$i = 1$ to $r$}
            \STATE $\hat{\mathbf{q}_{\timestep}}, \hat{\mathbf{K}_{\timestep}} \gets$Apply RoPE($\mathbf{q}_{\timestep}, \mathbf{K}_{\timestep}, \mathcal{I}$)%
            \STATE $\text{s} \gets$ attn\_score($\hat{\mathbf{q}_{\timestep}}, \hat{\mathbf{K}_{\timestep}}$)
            \STATE $\mathbf{q}_{\timestep}, \mathbf{K}_{\timestep} \gets$Apply RoPE($\mathbf{q}_{\timestep}, \mathbf{K}_{\timestep}$, $\mathcal{I}_{i-1}^{e}$)%
            \FOR{$j$ in $\mathcal{I}_{i-1}^{\hat{e}}$}
                \STATE $\mathbf{q}_{\timestep}', \mathbf{K}_{\timestep}' \gets$ Apply RoPE($\mathbf{q}_{\timestep}, \mathbf{K}_{\timestep}, j$)%
                \STATE $\text{s}' \gets$  attn\_score($\mathbf{q}_{\timestep}', \mathbf{K}_{\timestep}'$)%
                \STATE $\text{distance}[j] \gets \|\text{s} - \text{s}'\|_1$
            \ENDFOR
            \STATE $j^* \gets \arg\min (\text{distance})$
            \STATE Append $j^*$ to $e$
        \ENDFOR
        
        \RETURN $e$
    \end{algorithmic}
    \label{alg:RoPElite}
\end{algorithm}

%% file: figures_latex/performance_of_top-r.tex
\begin{figure}[htb]
\centering
\includegraphics[width=.95\linewidth]{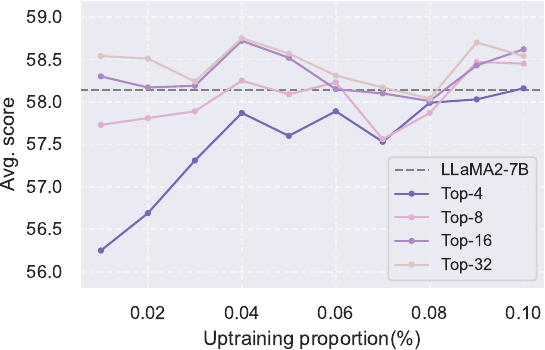}
\caption{
Performance of top-$r$ chunks. 
\textbf{Uptraining proportion} represents the proportion of tokens relative to the total number of tokens used during training the original model. 
}
\label{fig:performance of top-r}
\end{figure}

%% file: figures_latex/S-LRD_and_J-LRD.tex
\begin{figure*}[tb]
\centering
\includegraphics[width=.95\linewidth]{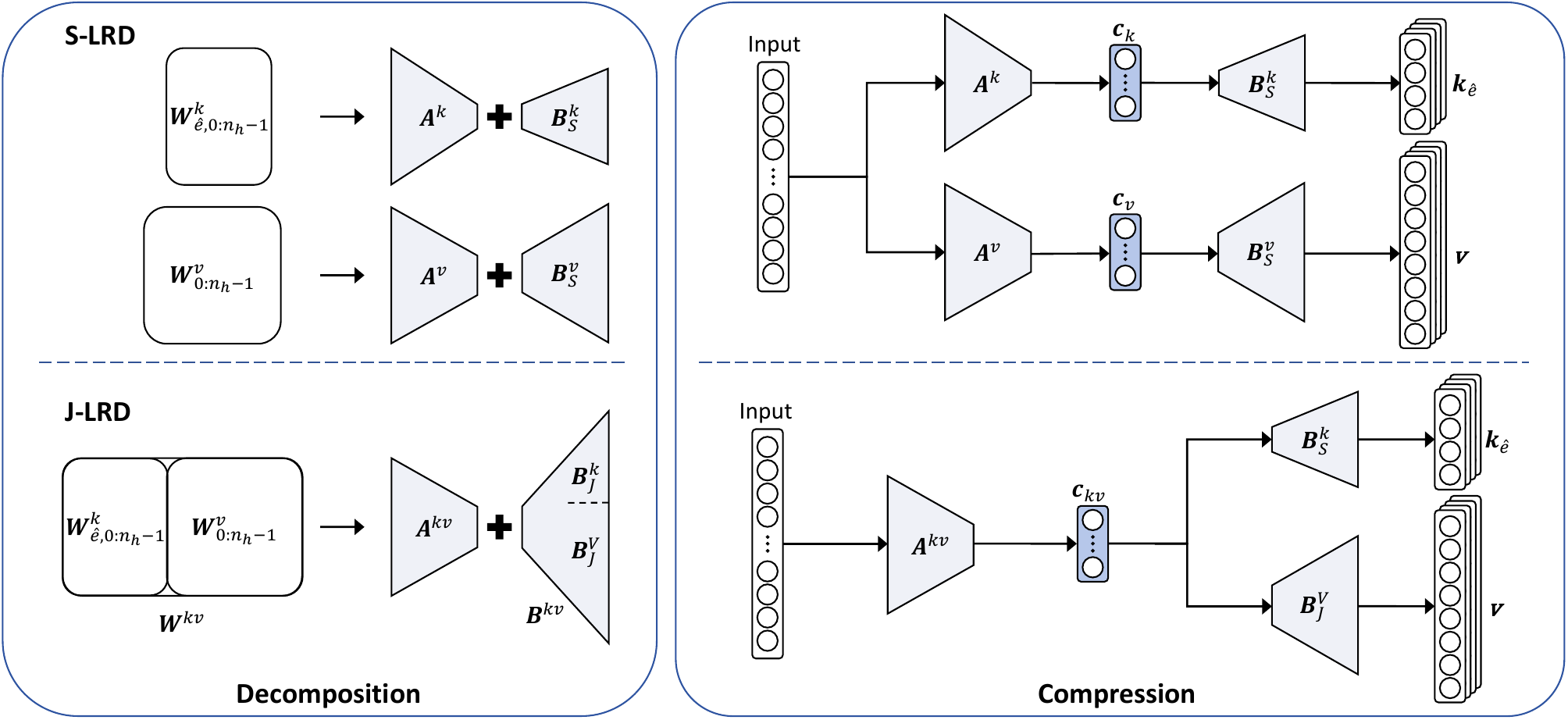}
\caption{
The projection matrices processed using S-LRD and J-LRD (Left), and the roles of the resulting matrices in the attention computation flow (Right).
}
\label{fig:different LRD}
\end{figure*}

%% file: sections/4_experiments.tex
\input{tables/overall_performance}

\section{Experiments}

\subsection{Benchmarks and Training Details}
\noindent\textbf{Models and Datasets.}
We conducted the primary experiments on the widely used open-source LLaMA2-7B \citep{DBLP:journals/corr/abs-2307-09288} model, and scaling experiments at the model size level on the LLaMA2-13B model.
The pretraining data, RefinedWeb \citep{DBLP:conf/nips/PenedoMHCACPAL23}, is open-source and of comparable quality to the original LLaMA2 pretraining data (something not achievable by the newer models), which is also the source of all the training data we used.

\noindent\textbf{Perplexity Evaluation.}
We extracted a portion of the data from RefinedWeb as a holdout dataset to calculate the language perplexity of the model.

\noindent\textbf{Benchmarks.}
To comprehensively evaluate the capabilities of the model, we assess its ability in commonsense reasoning, world knowledge, reading comprehension, and math reasoning. Specifically, we employ lm-evaluation-harness \citep{eval-harness} to assess its performance on 8 tasks, including BoolQ \citep{clark-etal-2019-boolq}, HellaSwag \citep{zellers-etal-2019-hellaswag}, OpenBookQA \citep{mihaylov-etal-2018-suit}, Winogrande \citep{DBLP:journals/cacm/SakaguchiBBC21}, GSM8K \citep{DBLP:journals/corr/abs-2110-14168}, TriviaQA \citep{joshi-etal-2017-triviaqa}, ARC easy and challenge \citep{DBLP:journals/corr/abs-1803-05457}. All tasks are evaluated under the zero-shot setting, except for GSM8K (8-shot) and TriviaQA (5-shot).

\noindent\textbf{Training Details.}
All models are trained with AdamW optimizer \citep{DBLP:conf/iclr/LoshchilovH19}, with $\beta=[0.9,0.95]$ and weight decay of 0.1.
We employ a constant learning rate, which is set to the minimum learning rate at the end of the model's pretraining phase.
All training was conducted on the NVIDIA H100 platform, using ZeRO-2 \citep{DBLP:conf/sc/RajbhandariRRH20} for memory management. Each training step includes 512 sequences, each with a length of 4096 tokens.

\subsection{Experimental Result}

We performe uptraining on the model with multiple target KV cache sizes, using less than $0.6\%$ of the tokens from the model's original pretraining phase. 
We compare \ourmethod{} with GQA \citep{ainslie-etal-2023-gqa}, a method that modifies the model structure to benefit from optimizations in the KV cache. 
The performance results on several benchmarks are shown in Table \ref{tab:main results}. 
\ourmethod{} outperforms across all tested KV cache ratios. 
As the KV cache size decreases, the slower performance degradation allows \ourmethod{} to retain performance comparable to the original model even when the KV cache is reduced to $25\%$. 
Additionally, thanks to the flexibility of the \ropemethod{} and J-LRD dimension combinations, \ourmethod{} is able to adjust the target KV cache ratio with finer granularity.

\subsection{Analysis and Discussion}

\subsubsection{Retention of Rotated Dimensions}
\label{sec:ablation rope}
It is essential to determine that trained RoPE-based models exhibit frequency preference differences at the head level. 
To address this, we designed a method that uniformly retains a specified number of rotated dimensions across frequencies, which called \textit{Uniform}.
Also, it is crucial to demonstrate the superiority of \ropemethod{} over other methods that determine rotated dimensions at the head level. 
We use the L2 norm of the RoPE chunks corresponding to each frequency as a measure of the contribution of each frequency’s corresponding dimension to the attention score. 
This measure is referred to as \textit{Contribution} \citep{hong-etal-2024-token, barbero2025round}.

\input{tables/rope_search}

We reduce the number of RoPE chunks in the model to a specified proportion using the two methods outlined above, as well as \ropemethod{}. The proportions tested ranged from $1/2$ to $1/16$. The model was then uptrained using less than $0.1\%$ of the original model's pretraining tokens.
As shown in Table \ref{tab:rope search}, \textit{Uniform} performs similarly to other methods at higher proportions. 
However, as the proportion decreases, the performance gap between it and the others that customize dimensions per head becomes increasingly apparent. 
Across all tested proportions, \ropemethod{} consistently outperform the other approaches, with the performance difference becoming more pronounced at lower proportions.
This further confirms the importance of identifying a dedicated frequency preference pattern for each attention head and demonstrates the effectiveness of the \ropemethod{} in the search for optimal rotated dimensions.

\subsubsection{Low-Rank Decomposition}
\label{sec:ablation LRD}

Section \ref{sec:LRD} provides an analysis of the parameter count and KV cache size for both S-LRD and J-LRD. 
It is evident that, for the same parameter constraints, J-LRD results in a smaller KV cache size compared to S-LRD.
Additionally, we used a greedy algorithm to find the optimal combination of $d_{c_k}$ and $d_{c_v}$ for S-LRD, given the KV cache size for each layer of attention. 
We then evaluate the language perplexity of both methods using the models trained in Section \ref{sec:ablation rope}. 

\input{figures_latex/LRD}

As shown in Figure \ref{fig:LRD}, in multiple settings, even though S-LRD has the advantage of customizability, it fails to outperform J-LRD when the KV cache size is fixed. 
This confirms the forward-looking nature of J-LRD, which leverages shared information between K and V.

\subsubsection{Impact of Training Token Quantity on Model Performance Recovery}
Since this method inevitably requires uptraining, a key concern is the speed at which model performance recovers at different KV cache ratios. 
As shown in Figure \ref{fig:performance of different ratio}, for higher KV cache sizes, the model performance converges quickly. 
However, for lower ratios, such as $12.5\%$, more training effort is required. 
Therefore, when aiming for a model with a KV cache size smaller than $25\%$ of the original, one must choose the cache size based on the acceptable training cost, or combine it with other techniques to reduce KV cache size, such as quantization \citep{DBLP:conf/mlsys/ZhaoLZYC0CK0K24, DBLP:conf/nips/HooperKMMSKG24, DBLP:conf/icml/LiuYJZXBC024}.

\input{figures_latex/performance_of_different_ratio}

\subsubsection{Performance Differences Across Model Sizes}

In general, larger models have more attention heads, which means that when the same number of chunks is retained for each head, the multi-head attention mechanism aggregates more positional information. 
Additionally, larger weight matrices may also contain more redundant information.  
We explored the larger LLaMA2-13B model and used the percentage of performance loss during training as an observation point for training dynamics.  
As shown in Figure \ref{fig:model size}, when uptraining the same number of tokens, larger models exhibit faster convergence. 
The relative performance loss across models of different scales reaches a similar upper bound, suggesting that the performance loss due to convergence limits is consistent across model sizes.

\input{figures_latex/model_size}

%% file: tables/overall_performance.tex
\begin{table*}[!t]
\centering
\resizebox{\linewidth}{!}{
\begin{tabular}{rlccccccrccc}
\hline
    \textbf{Cache} & \textbf{Method} & \textbf{Arc-C} & \textbf{Arc-E} & \textbf{BoolQ} & \textbf{HS} & \textbf{OB} & \textbf{WG} & \textbf{GSM} & \textbf{TQA} & \textbf{Avg.(6)} & \textbf{Avg.(8)}\\ 
\hline
    100.0 & LLaMA2-7B & 46.08 & 74.54 & 77.77 & 75.84 & 44.00 & 68.75 & 14.18 & 63.99 & 64.50 & 58.14  \\
\hline
    \multirow{2}{*}{50.0} & GQA & 43.94 & 72.47 & 73.98 & 73.76 & 41.60 & 68.43 & 8.04 & 60.39 & 62.36 & 55.33  \\
     & \ourmethod{} & 47.10 & 74.92 & 75.63 & 75.66 & 44.00 & 69.22 & 11.22 & 64.02 & \textbf{64.42} & \textbf{57.72}  \\
\hline
    34.4 & \ourmethod{} & 46.33 & 74.75 & 77.03 & 75.51 & 43.40 & 69.06 & 11.75 & 63.38 & \textbf{64.33} & \textbf{57.65}  \\
\hline
    28.1 & \ourmethod{} & 45.99 & 74.28 & 77.25 & 75.38 & 44.40 & 69.53 & 11.83 & 63.05 & \textbf{64.47} & \textbf{57.71}  \\
\hline
    \multirow{2}{*}{25.0} & GQA & 40.78 & 70.37 & 72.23 & 71.81 & 40.80 & 66.14 & 3.11 & 55.51 & 60.36 & 52.59  \\
     & \ourmethod{} & 44.88 & 74.49 & 76.51 & 75.21 & 43.20 & 69.06 & 12.51 & 62.51 & \textbf{63.89} & \textbf{57.30}  \\
\hline
    21.9 & \ourmethod{} & 44.11 & 73.40 & 74.86 & 74.94 & 43.00 & 68.82 & 10.39 & 62.00 & \textbf{63.19} &\textbf{56.44}  \\
\hline
    \multirow{2}{*}{12.5} & GQA & 40.10 & 69.11 & 68.26 & 69.84 & 39.80 & 62.83 & 2.05 & 50.49 & 58.32 & 50.31  \\
     & \ourmethod{} & 44.20 & 73.99 & 74.46 & 74.40 & 41.40 & 67.56 & 9.40 & 59.94 & \textbf{62.67} & \textbf{55.67}  \\
\hline
\end{tabular}
}
\caption{
The scores of \ourmethod{} and GQA on 8 benchmarks.
Avg.(6) for the first 6 and Avg.(8) represents all.
\textbf{Cache} represents the proportion of the current KV cache relative to that of the original model. 
}
\label{tab:main results}
\end{table*}

%% file: tables/rope_search.tex
\begin{table}[htb]
\centering
\resizebox{\linewidth}{!}{
\begin{tabular}{|l|c|c|c|c|}
\hline
    \textbf{Method} & \textbf{$r=32$} & \textbf{$r=16$} & \textbf{$r=8$} & \textbf{$r=4$} \\ \hline
    \textit{Uniform} & 58.33 & 58.11 & 57.69 & 57.49 \\ 
    \textit{Contribution} & 58.24 & 58.30 & 58.00 & 57.84 \\ 
    \ropemethod{} & \textbf{58.54} & \textbf{58.62} & \textbf{58.45} & \textbf{58.16} \\ \hline
\end{tabular}
}
\caption{
Average scores of different rotation dimension search methods across $8$ benchmarks. 
$r$ represents the number of 2D chunks assigned to each attention head, responsible for a specific frequency.
When $r = 64$, corresponding to the original model, the performance is 58.14.
}
\label{tab:rope search}
\end{table}

%% file: figures_latex/LRD.tex
\begin{figure}[htb]
\centering
\includegraphics[width=\linewidth]{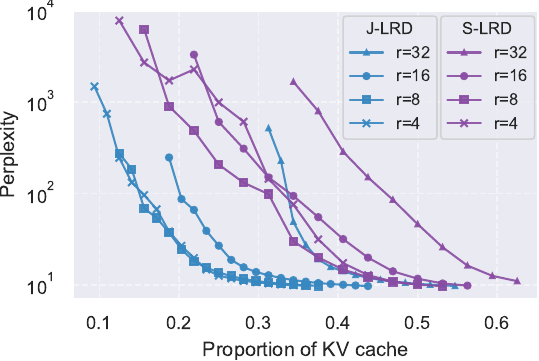}
\caption{
The perplexity of the \ropemethod{} model on the dataset with varying compression ratios of the KV cache, as the number of frequency-related chunks retained for each attention head changes under S-LRD and J-LRD. 
Only points without additional parameter overhead are shown.
}
\label{fig:LRD}
\end{figure}

%% file: figures_latex/performance_of_different_ratio.tex
\begin{figure}[htb]
\centering
\includegraphics[width=\linewidth]{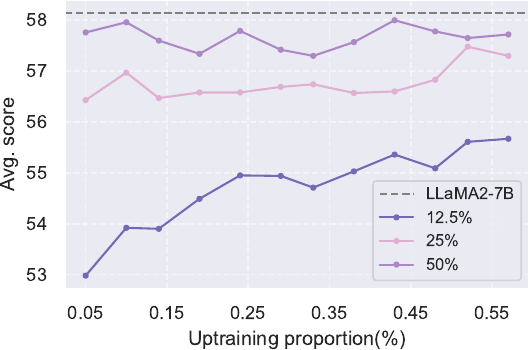}
\caption{
The performance trend of the model after applying \ourmethod{}, as training progresses, under varying KV cache compression ratios.
}
\label{fig:performance of different ratio}
\end{figure}

%% file: figures_latex/model_size.tex
\begin{figure}[htb]
\centering
\includegraphics[width=\linewidth]{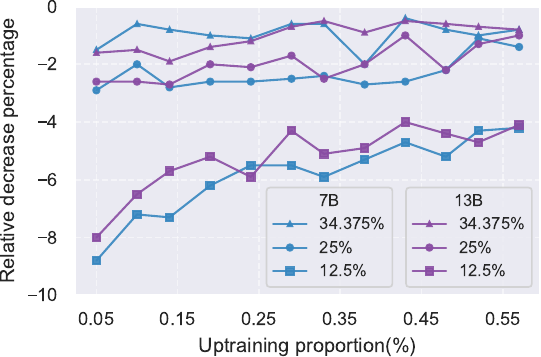}
\caption{
The relative percentage decrease in average scores across $8$ benchmarks for LLaMA2 family models (7B and 13B) after applying \ourmethod{}, under different KV cache compression ratios.
}
\label{fig:model size}
\end{figure}

%% file: sections/5_related.tex
\section{Related Work}
\label{sec:related work}

\noindent\textbf{SVD for KV Cache Compression}
Many works have focused on performing low-rank decomposition of the key and value projection matrices into $\mathbf{A}^{k}\mathbf{B}^{k}$ and $\mathbf{A}^{v}\mathbf{B}^{v}$ \citep{DBLP:journals/corr/abs-2410-03111, DBLP:journals/corr/abs-2410-21465, chang2025palu}. 
Like these methods, our approach also reduces the intermediate tensor dimensions by absorbing $\mathbf{B}^{k}$ and $\mathbf{B}^{v}$ into the rest of the matrix computations to cache lower-dimensional intermediate tensors \citep{DBLP:journals/corr/abs-2405-04434, DBLP:journals/corr/abs-2412-19437}. 
However, because of the non-linear nature of positional embeddings, these methods can only cache $\mathbf{x}\mathbf{A}^{k}$ and $\mathbf{x}\mathbf{A}^{v}$ as the new KV cache. 
This requires designing specialized GPU kernels to reconstruct the rotation process of the keys at each decoding step.
Thanks to \ropemethod{}, our approach achieves KV cache efficiency not by handling the non-linear components but by leveraging the remaining linear parts.

\noindent\textbf{Efficient MHA Variants for KV Cache}
Several notable efforts have aimed at reducing the use of KV cache through the redesign of multi-head attention mechanisms. 
GQA \citep{ainslie-etal-2023-gqa} and MQA \citep{DBLP:journals/corr/abs-1911-02150} (GQA with group size 1) focus on reducing the number of K and V heads. MLA \citep{DBLP:journals/corr/abs-2405-04434, DBLP:journals/corr/abs-2412-19437} and MFA \citep{DBLP:journals/corr/abs-2412-19255} aim to project K and V into a lower-rank space.
In contrast to our approach, which operates on pretrained models, these methods are applied at the model design stage before pretraining, with only GQA allowing modifications based on an already trained model. 
Our method, by principle, can be applied to both pretrained MHA-based models and QGA-based models, and the modified attention structure can also inform future model design.

%% file: sections/6_conclusion.tex
\section{Conclusion}
We propose \ourmethod{} for reducing the KV cache size in RoPE-based models with high flexibility, composed of two main components.
First, \ropemethod{} identifies the varying frequency preferences across different attention heads in RoPE-based models, demonstrating that, after a small amount of uptraining, retaining only the most important dimensions of RoPE still allows the model to recover its original performance. 
This enables subsequent low-rank compression of the KV cache to avoid dealing with the non-linear components of RoPE.
Second, for the compression part, J-LRD takes into account the shared information between the K and V projection matrices during their decomposition.
This allows the KV cache to be further reduced compared to traditional methods that treat the projection matrices separately.
Experimental results show that, compared to GQA, \ourmethod{} recovers performance similar to the original model more quickly under higher KV cache settings. 
In lower KV cache settings, the relative performance loss is less than $1/3$ of that observed in the baseline.

%% file: sections/8_limitations.tex
\section*{Limitations}
Since pretrained models with publicly available datasets are becoming increasingly rare, and as the number of tokens used in the pretraining phase of base models grows (e.g., 15T for LLaMA3), foundation models are becoming increasingly well-trained. 
This leads to a higher demand for tokens during modifications to restore model performance. 
Therefore, the models explored in this paper are limited.

%% file: sections/7_appendix.tex
\section{More Detail on the Preference Pattern of Each Attentional Head}
\label{sec:appendix pattern}

For a more detailed frequency preference pattern, refer to Figure \ref{fig:appendix top-r}.

\input{figures_latex/appendix_top-r}

\section{Analysis of \ropemethod{} Time Consumption}
\label{sec:appendix time comsumption}

We analyze the time complexity of RoPElite. 
Our method computes the top-$r$ chunks for each layer and head with a time complexity of $O(rh_d)$ forward pass. To obtain the top-$r$ chunks, the algorithm requires $r$ iterations. In each iteration, the influence of each chunk in $\mathcal{I}_{r}^{\hat{e}}$ on the attention scores is computed, and the chunk that most closely approximates the original RoPE attention scores is selected and added to $\mathcal{I}_{r}^{e}$.

In practice, the computation for different heads is parallelized. Although the top-$r$ selection differs across heads, the calculations of attention score are independent for each head. This allows us to leverage the parallelism of matrix operations, computing the influence of a given chunk on attention scores for all heads in a single forward pass. 

Calculations for different layers can also be performed simultaneously within the same forward pass. During the computation, we only calculate the attention scores of applying RoPE on specific chunks within the attention block, while using the attention scores of the original RoPE during the forward pass. This approach ensures the independence of the algorithm across different layers.

\section{Dimension Allocation of \ropemethod{} and LRD}

For a given KV cache size, there exist numerous valid dimension configurations for both \ropemethod{} and the LRD components. 
This introduces an optimization challenge, as selecting the appropriate dimensions of \ropemethod{} and LRD to achieve the optimal performance requires balancing the contributions of both components under the constraint of a fixed total size.  
To simplify this allocation problem, we adopt the following strategies.

\noindent\textbf{Hardware-Friendly.}
For optimal hardware compatibility, the primary design principle is to ensure that the intermediate dimension of LRD, $c_{kv}$, is aligned with multiples of 128 across all valid configurations.
\footnote{For further details on the requirements and performance considerations of Tensor Cores, refer to \href{https://docs.nvidia.com/deeplearning/performance/dl-performance-matrix-multiplication/index.html\#requirements-tc}{NVIDIA's official documentation}.}

\noindent\textbf{No Additional Parameters.}
Since low rank decomposition may lead to an increase in the number of parameters, a fair approach to prevent sacrificing parameter efficiency for a smaller KV cache is to ensure that the total number of parameters does not increase after applying low rank decomposition.

\noindent\textbf{Lower Perplexity.}
Under these two constraints, a set of potential dimension configurations is selected. 
These configurations are then evaluated based on language perplexity on a holdout dataset, and those with lower perplexity are chosen as the final dimension design for a given KV cache.

%% file: figures_latex/appendix_top-r.tex
\begin{figure*}[htb]
\centering
\includegraphics[width=\linewidth]{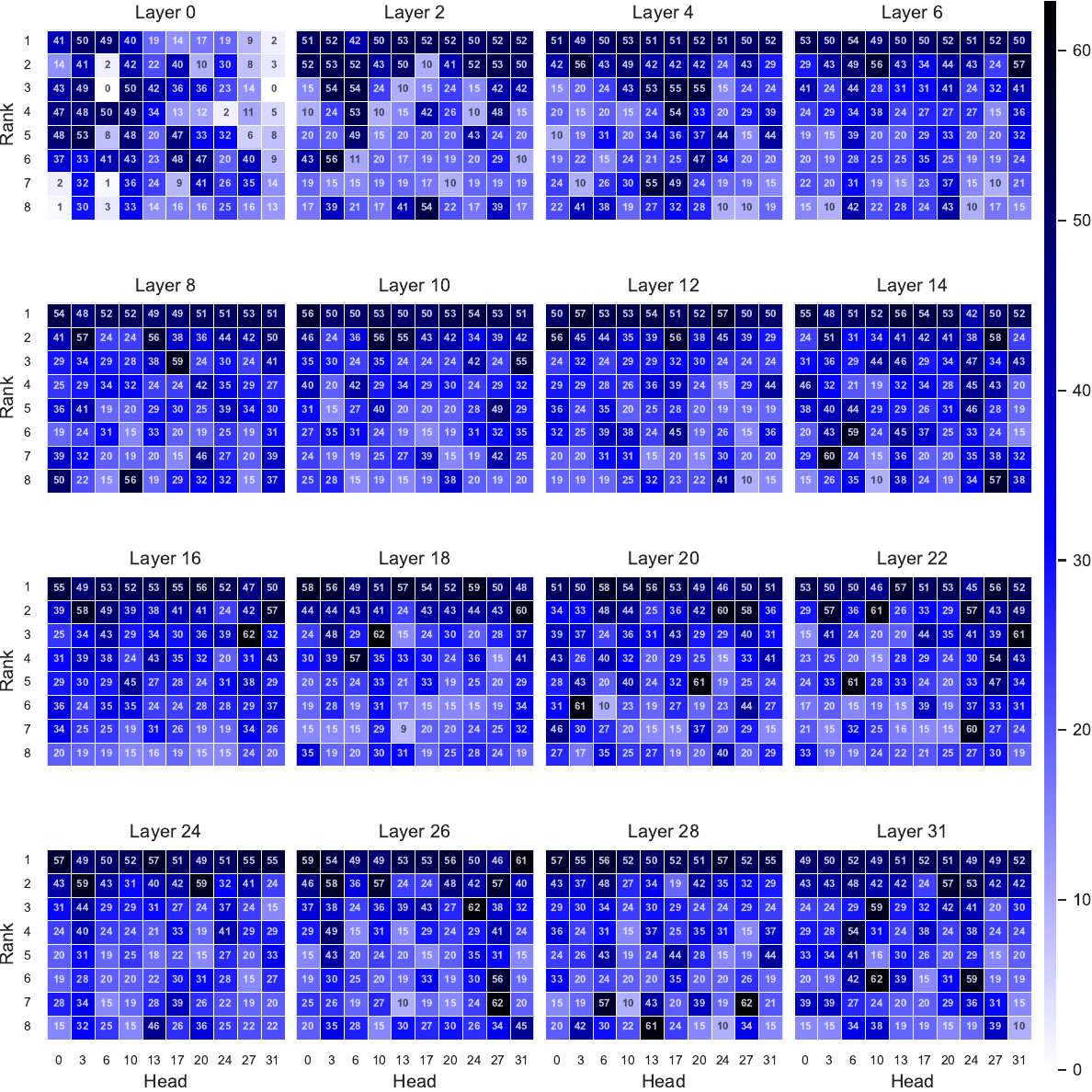}
\caption{
Top-$r$ chunks of different attention heads in different layers. 
}
\label{fig:appendix top-r}
\end{figure*}